\documentclass[10pt,twocolumn,letterpaper]{article}

\usepackage{wacv}
\usepackage{times}
\usepackage{epsfig}
\usepackage{graphicx}
\usepackage{amsmath}
\usepackage{amssymb}
\usepackage{booktabs}       % professional-quality tables
\usepackage{subcaption}
\makeatletter
\@namedef{ver@everyshi.sty}{}
\makeatother
\usepackage{tikz}
\usepackage{pgfplots}
\usepackage{multirow}
\usepackage[normalem]{ulem}

\usepackage{pifont}% http://ctan.org/pkg/pifont
\newcommand{\cmark}{\ding{51}}%
\newcommand{\xmark}{\ding{55}}

\def\wacvPaperID{1213} % Enter the WACV Paper ID here

\wacvfinalcopy % *** Uncomment this line for the final submission

\ifwacvfinal
\def\assignedStartPage{1} % *** Enter the assigned starting page number (instead of 9876)
\fi

\ifwacvfinal
\usepackage[breaklinks=true,bookmarks=false]{hyperref}
\else
\usepackage[pagebackref=true,breaklinks=true,colorlinks,bookmarks=false]{hyperref}
\fi

\ifwacvfinal
\else
\fi

\begin{document}

%%%%%%%%% TITLE
\title{Rescaling CNN through Learnable Repetition of Network Parameters}

\author{Arnav Chavan$^{\ast}$, Udbhav Bamba$^{\ast}$, Rishabh Tiwari$^{\ast}$ and Deepak K. Gupta\thanks{All authors contributed equally.}\\
Transmute AI Lab, Texmin Hub, IIT Dhanbad, India \\
{\tt\small \{arnav,udbhav,rishabh,deepak\}@transmute.ai}
}

% \author{Arnav Chavan\\
% {\tt\small arnavchavan04@gmail.com}
% \and
% Udbhav Bamba\\
% {\tt\small ubamba98@gmail.com}
% \and
% Rishabh Tiwari\\
% {\tt\small akchitra99@gmail.com}
% \and
% Deepak Gupta\\
% {\tt\small guptadeepak2806@gmail.com} \\
% Transmute AI Lab, Texmin Hub, IIT Dhanbad, India\\
% }

\maketitle
%\thispagestyle{empty}

%%%%%%%%% ABSTRACT
\begin{abstract}
   Deeper and wider CNNs are known to provide improved performance for deep learning tasks. However, most such networks have poor performance gain per parameter increase. In this paper, we investigate whether the gain observed in deeper models is purely due to the addition of more optimization parameters or whether the physical size of the network as well plays a role. Further, we present a novel rescaling strategy for CNNs based on learnable repetition of its parameters. Based on this strategy, we rescale CNNs without changing their parameter count, and show that learnable sharing of weights itself can provide significant boost in the performance of any given model without changing its parameter count. We show that small base networks when rescaled, can provide performance comparable to deeper networks with significantly reduced model size.

Our repetition strategy scales well for large datasets (\emph{e.g.}, ImageNet). Moreover, it tends to learn important features that help the model to generalize well across tasks. For example, backbone network of a RepeatNet model scaled on the task of image classification is shown to work well as an encoder for the task of segmentation. The relevance of learnable repetitions weight sharing is further highlighted through the example of group-equivariant CNNs. We show that the significant improvements obtained with group-equivariant CNNs over the regular CNNs on classification problems could only partly due to the added equivariance property, and part of it comes from the learnable repetition of network weights. For Rot-MNIST dataset, we show that up to 40\% of the relative gain reported by state-of-the-art methods for rotation equivariance could actually be due to just the learnt repetition of weights.%\footnote{Code is publicly available at https://github.com/transmuteAI/RepeatNet}
\end{abstract}

%%%%%%%%% BODY TEXT
\section{Introduction}
One of the major reasons for the popularity of deep learning-based models is their ability to easily adapt to different problems, ranging from simple input to output mapping to those involving extremely complex underlying distributions. A recent popular strategy to improve such models is to employ a base model and scale it in size, thereby improving its discriminative power. Due to largely increased number of parameters, such approaches have been able to achieve state-of-the-art (SOTA) performance on popular benchmark datasets such as CIFAR10/100 \cite{Krizhevsky2009LearningML} and ImageNet \cite{imagenet_cvpr09}.

A common way to scale up CNNs is to make its layers wider or the network itself deeper. For example, VGG \cite{simonyan2014very} increases the model depth through stacking multiple layers together. However, vanishing gradients \cite{GlorotAISTATS2010} harm the performance of such stacked models due to which going deeper beyond a certain limit produces adverse affects on performance. Residual Networks (ResNets) \cite{he2016deep} overcome the outlined issue through adding residual connections for better gradient flow, thus enabling the construction of even deeper models. Wide ResNets \cite{zagoruyko2016wide} take the orthogonal direction of making the layers wider through scaling the number of channels in each convolutional layer by a variable parameter. This approach brings in significant boost in the performance of the models. Most current SOTA models in the field of computer vision now use compound scaling which involves uniformly scaling up the width, depth and resolution of the \mbox{models \cite{tan2019efficientnet}}.

It is evident from the developments outlined above that the deeper and wider models exhibit stronger discriminative power and can better extract the rich feature representations contained in images. Thus
% {However, most such scaled up models have a poor performance gain per parameter increase. For example, ResNet101 has $\sim$75\% more parameters as compared to ResNet50 for a mere $\sim$1\% performance gain on the ImageNet dataset. Similar pattern can also be observed for the recent EfficientNet models where going from EfficientNet B4 to B7 increases parameters by $\sim$250\% for a $\sim$1.5\% performance improvement. Such adjustments lead to drastic increase in the memory requirements and it is important that these are efficiently utilized by gaining a better insight on their impact.  Along this line,} 
it is of major interest to identify whether such gains are actually due to the increased optimization parameters or whether part of it is due to the physical size of the scaled model. The answer being latter would imply that if we can design a mechanism to upscale the model, we might be able to improve its performance to some extent even without the need of increased optimization parameters.
%we do not necessarily need to increase the optimization parameters, but only devise a mechanism to scale the size of the model.
In this paper, we investigate the importance of network scaling at fixed parameter count. To this end, we first demonstrate that a well-drafted scaling policy of an existing architecture can already improve its performance for the same number of optimization parameters. Referred henceforth as `RepeatNet', our scaling strategy is based on the principle that transformed version of the learned weights from one part of the network, when used again in another part, results in scaled-up architectures that improve the model performance. For example, a VGG4 model trained for CIFAR-100 dataset, when scaled up using the simplest linear variant of RepeatNet to match the physical size of VGG11, shows a massive absolute improvement of approximately 19\%.

We explore two ways of scaling the network architecture: \emph{depth scaling} and \emph{width scaling} by repeating different projections of same set of filters. While the linear projections can be used to effectively scale the depth of an existing network, it is degenerate for width scaling (later discussed in Section \ref{sec:linear_projections}). For more efficient and universal scaling of the existing CNN models, we further propose two nonlinear scaling strategies ($s$-RepeatNet and $f$-RepeatNet) that overcome the limitations of the linear scaling. 
% \udbhavr{}{We propose novel ways of weight repetition which involves using different projections of same set of filters.} 
Our $s$-RepeatNet approach draws inspiration from the way nonlinear activation functions improve the representation capability of the networks. It adapts the popular swish activation function \cite{ramachandran2017searching} to create transformed variants of the learned weights that can be effectively deployed in other parts of the network. The $f$-RepeatNet variant learns sign flipping of the repeated weights, with optimization objective similar to BiRealNet \cite{bireal} to increase the representation capability of the scaled models.

The efficacy of RepeatNet is demonstrated on several popular CNN architectures and benchmark datasets. We show that RepeatNet provides significant boost in performance with negligible increase of network parameters compared to the respective base models. Another important aspect of deep learning models is their transferability across different tasks. In this regard, we also train scaled-up CNN architectures on ImageNet dataset \cite{imagenet_cvpr09} for the task of classification, and show that the trained backbone performs well for the downstream tasks. We show that pre-trained scaled up versions of ResNet-18 model when used as an encoder for the segmentation model, achieves an increase of more than 5\% in the mIOU metric on PASCAL-VOC dataset \cite{Everingham10}.
% {task of segmentation. For ResNet-18 model, we show that when scaled up by a factor of 4 on the task of classification of ImageNet samples, the resultant model, used as an encoder for the segmentation model, achieves an increase of more than 5\% in the mIOU metric on PASCAL-VOC dataset.}

We further demonstrate using a popular example from the field of deep learning (rotation-equivariant CNN or RE-CNN \cite{e2cnn}) that ignoring the impact of repetition of weights can lead to erroneous conclusions when comparing different models. While several variants exists, we focus on the RE-CNN model presented in \cite{e2cnn}. Their strategy uses steerable filters \cite{cohen2016steerable} that compute orientation dependent responses for multiple orientations efficiently through sharing of weights \cite{weiler2018learning, e2cnn}. It is very common to compare the performance of RE-CNNs and other similar models with regular CNNs for the same number of optimization parameters. However, since RE-CNNs employ weight sharing, the realized parameter set and eventually the FLOPs are higher during inference time when compared with an equivalent CNN model. This raises the question whether the gain in performance observed with RE-CNN on datasets such as Rot-MNIST \cite{larochelle2007empirical} is purely due to the induced property of rotation equivariance of the model, or whether part of it is due to repetition of weights. We address this concern later in the paper.

To summarize, the contributions of this paper can be outlined as follows.

\begin{itemize}
    \item We demonstrate through numerical experiments that the gain observed in model performance due to the use of increased number of optimization weights could partly be accounted to scaling up of the physical size of the network itself.
    \item For efficient sharing of weights across the network, we present \emph{RepeatNet}, a learnable filter transformation strategy that can be used to effectively scale CNNs along the width and depth dimensions. For increased flexibility of representations in weight sharing, we present two nonlinear transformation strategies ($s$-RepeatNet and $f$-RepeatNet).
    \item We demonstrate through experiments on popular CNN architectures and datasets (including ImageNet) that RepeatNet-scaled architectures achieve significant boost in performance. Moreover, we demonstrate that features obtained from RepeatNet generalize well across tasks - the backbone of a RepeatNet architecture designed for classification can also be used for segmentation tasks.
    \item Understanding the influence of weight sharing in a network is important, and we demonstrate through a popular example that ignoring the impact of network scaling can lead to erroneous conclusions in deep learning studies.
\end{itemize}

\section{Related Works}
Scaling the network is one among the most popular ways to improve the performance of CNN models. Several previous works have addressed this aspect in the past. For example, VGG architectures \cite{simonyan2014very} are scaled by increasing the depth of the architecture. Another class of architectures is WideResNet \cite{zagoruyko2016wide} that are scaled by making the network wider. Recent works have shown that a more efficient strategy to scale up a network is to grow along depth as well as its width. Referred as EfficientNets \cite{tan2019efficientnet}, such models use a mix of depth, width and resolution scaling, and are shown to have a much better performance per parameter ratio in comparison to ResNets.

The interplay of parameter and model physical size in model's performance is also visible in the field of pruning. For deep learning models, several works have shown that most often the networks are overparameterized, and it is possible to achieve similar performance with reduced number of parameters \cite{liu2017learning, frankle2018lottery, tiwari2021chipnet, lemaire2019structured, dai2018compressing}. A sub-class of pruning, referred as unstructured pruning reduces the number of parameters required to define an architecture, but it still preserves the model size \cite{srinivas2015data, han2015learning, molchanov2017variational, louizos2017learning}. Unstructured pruning strategies have been shown to work well, which also indicates that beyond the number of parameters, the physical size of the model as well plays role in controlling its performance. However, a clear understanding on this aspect is still missing. Nevertheless, such approaches do not harness the improvements brought in through repetition of the parameters. These approaches are orthogonal to our repetition strategy, and we believe that these could be combined with our repetition approach presented in this paper.

Scaling up a network through weight sharing is another popular approach that is frequently adopted. ShaResNet shares the weights of convolutional layers between residual blocks in ResNets operating at the same spatial scale \cite{boulch2017sharesnet}. Similarly, DACNNs share the same kernel weights across layers using additional $1\times 1$ convolutional layers, thereby reducing the overall model parameters \cite{huang2019deep}. While these works present the general concept of weight sharing, they either don't consider the correlation between shared parameters, require additional layers to counter it or are limited to sharing at same spatial scales. Weight sharing is also observed in several variants of group-equivariant networks \cite{cohen2016steerable, weiler2018learning}. While these networks focus on making network architectures robust against certain transformations, an implicit network scaling occurs due to multiple filters blocks used to account for several orientations. The methods generally ignore the affect of network scaling, and we show in this paper that part of their improvement could solely be due to network scaling.

\begin{figure*}[h]
\begin{center}
\begin{tikzpicture}
\node[inner sep=0pt] (russell) at (0,0)
    {\includegraphics[scale=0.7]{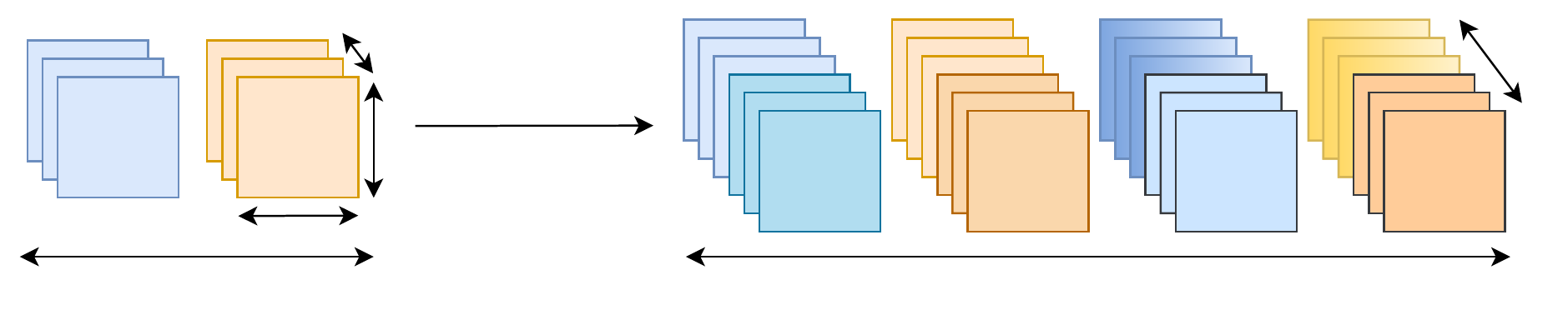}};
\node[] at (-5.0,-1.1) {{\small $n$}};
\node[] at (-3.4,1.1) {{\small $m$}};
\node[] at (-3.3,0.3) {{\small $k$}};
\node[] at (-4.1,-0.65) {{\small $k$}};
\node[] at (3.2,-1.1) {{\small $n \times \gamma_2$}};
\node[] at (-2.2,0.6) {{\small $\mathcal{\psi_{\boldsymbol\beta}(\cdot)}$}};
\node[] at (6.5,1.1) {{\small $m \times \gamma_1$}};
\end{tikzpicture}
\vspace{-1em}
\caption{Schematic representation of our RepeatNet strategy showing the generation of child filter sets (right) from the parent filter sets (left).}
\label{fig_schem1}
\end{center}
\end{figure*}
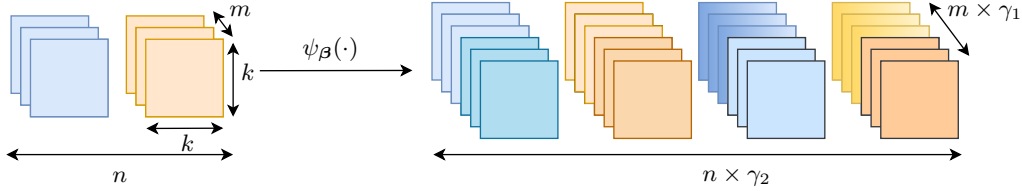

\section{Methodology}
In this section, we first present the core principles of the trainable filter repetitions. We then explain the procedure to scale any neural network based on this principle.

%%%%%%%%%%%%%%%% FILTER REPETITION %%%%%%%%%%%%%%%%%%
\subsection{Filter Repetition}
In simple terms, our filter repetition approach involves taking a \emph{parent} filter set from a particular convolutional layer and transforming it to different \emph{child} sets through multiple repetitions across the input and output channel dimensions. Same child filter used at different parts of the network can have adverse effects on the performance, especially because of the varying receptive field at different depths. To circumvent this issue and improve the learning capability of the model, each child filter set can be passed through a learnable transformation function. 

Let $\mathcal{F} \in \mathbb{R}^{k \times k \times m \times n}$ denote a filter block, comprising a set of convolutional filters of size $k \times k$, where $m$ and $n$ correspond to the number of input and output channels, respectively. We define filter repetition as the mapping \mbox{$\mathcal{F}_p \xrightarrow{\mathcal{\psi_{\boldsymbol\beta}(\cdot)}} \mathcal{F}_c$}, where $\mathcal{F}_p$ and $\mathcal{F}_c$ denote the parent and child filter blocks, respectively, and $\psi_{\boldsymbol\beta}(\cdot)$ is a learnable nonlinear transformation function. In this setting, the weights in the parent block are trainable and learnt during the training process. The weights of the child block $\mathcal{F}_c$ depend on $\mathcal{F}_p$ as well as an additional parameter set $\boldsymbol\beta$. Note that $\boldsymbol\beta$ is generally a very small set of parameters or a set of boolean variables, and the added memory burden is negligible. For example, for one of the configurations proposed in this paper, length of this parameter set is 2 for every mapping.

Fig. \ref{fig_schem1} provides a schematic representation of how these repetitions are performed. Here, the trainable filter block $\mathcal{F}_p \in \mathbb{R}^{k \times k \times m \times n}$ is mapped to another larger block of dimensions \mbox{$k \times k \times \tilde{m} \times \tilde{n}$}. We propose to define network architectures such that $\tilde{m}$ and $\tilde{n}$ are integer multiples of $m$ and $n$, respectively. We define these multiplicities as $\gamma_1=\frac{\tilde{m}}{m}$ and $\gamma_2 = \frac{\tilde{n}}{n}$, thereby implying \mbox{$\gamma_1 \times \gamma_2$} repetitions in \mbox{Fig. \ref{fig_schem1}}. Let $\mathcal{F}_{c,i}$ denote the $i^{\text{th}}$ child filter block to be mapped with one of the repetitions. In simple terms, the goal is then to fill the entire set of weights in $\mathcal{F}_{c,i}$ using a dedicated parameter set $\boldsymbol\beta_i$ and the parent set  $\mathcal{F}_{p}$. 

Since the diversity across a set of child blocks obtained from a single parent depends heavily on the respective $\boldsymbol\beta$ set, the choice of this mapping function needs to be made cautiously. We show later in this paper that with a proper choice of $\psi_{\boldsymbol\beta}$, a large network (\emph{e.g.}, VGG11) can be expressed using only a very small fraction of its parameters at the expense of limited drop in performance of the model.

%%%%%%%%%%%%%%%% NETWORK SCALING %%%%%%%%%%%%%%%%%%
\subsection{Network Scaling}
Repetition of weights can be used to scale the existing CNN networks in two different ways: \emph{width scaling} and \emph{depth scaling}. Details related to these are provided below.

\textbf{Width scaling. }This scaling strategy involves increasing the number of filters through performing repetitions within the same convolutional layer. This is equivalent to having a smaller parent filter set, repeating it multiple times and concatenating all the repetitions to form the filters for a certain convolutional layer. This effectively allows a smaller convolutional layer to take in a larger number of input channels and return a larger number of output channels at the same time. Networks scaled this way are wider than their original counterparts with almost the same number of parameters.

\textbf{Depth scaling. }In this strategy, repetitions are performed across different convolutional layers of the network. An entire convolutional layer constitutes the parent set and it is repeated to generated child sets where each child filter set itself forms a different convolutional layer of the network. Parent filter set is itself a convolutional layer which is repeated accordingly to form multiple child set, each of which is then used as a convolutional layer of the network. Networks scaled through this strategy are much deeper than their original counterparts with almost the same number of parameters which effectively increases the receptive field of the network.

%%%%%%%%%%%%%%%% LEARNABLE FILTER TRANSFORMATION %%%%%%%%%%%%%%%%%%

\subsection{Learnable Filter Transformation}
\label{sec:linear_projections}
\textbf{Linear projections. }A simple yet effective way of scaling the existing networks is to directly use the parents filters at multiple child locations, thereby setting $\mathcal{F}_{c,i} = \mathcal{F}_p$. However, the scope of this strategy is limited to depth scaling, and it cannot be used to make networks wider. 

In general, expressing an entire filter block $\mathcal{F}_c$ with another parent block $\mathcal{F}_p$ is expected to limit the representation capability of the resultant weights. However, due to the nonlinear activations applied on the feature output at very hidden layer, the linear projects are expected to help in depth scaling of the network to some extent. We show later through experiments that for some cases, even the linear scaling boosts model performance significantly.

However, linear scaling cannot be used to effectively make networks wider. Repeating the weights within the same filter block is equivalent to concatenating the output feature map with its exact copies, thereby adding no useful information. 

\textbf{Nonlinear filter transformation. }To overcome the limitation of linear repetition, we employ learnable transformations functions that remove the correlation between the filter sets and increase the diversity across the filter sets obtained from repetition. We propose two different ways to build learnable nonlinear repetitions:

\emph{$s$-RepeatNet. }This strategy relies on a nonlinear projection that differentiates the outputs of two different child filter sets. The transformation function $\mathcal{\psi_{\boldsymbol\beta}(\cdot)}$ is defined as
%\dpk{phi should be used instead of psi}
\begin{equation}
    \psi_{\boldsymbol\beta}(x) = \frac{\beta_1 x}{1+e^{\beta_2 x}},
\end{equation}
where $\boldsymbol\beta = \{\beta_1, \beta_2\}$ is the pair of additional parameters to be learnt for each repetition. Further, the term $x$ is used to denote every element of the parent filter set $\mathcal{F}_p$. 
The above transformation function is inspired by the swish activation function \cite{ramachandran2017searching}, with the addition of the extra parameter $\beta_1$. 
Without this parameter, only a very limited set of projections can be learnt. Adding $\beta_1$ alleviates this restriction and adds extra flexibility for the repeated kernels to independently scale themselves. The parameter set $\boldsymbol\beta$ is also treated as the normal trainable parameters of the network similar to the filters, and these are also updated iteratively through back-propagation.

As outlined earlier, a total of $\gamma_1 \times \gamma_2$ repetitions of a parent filter set occur. This would lead to the negligible addition of only $2\times \gamma_1 \times \gamma_2$ extra training parameters.

\emph{$f$-RepeatNet. }This strategy is similar to the linear projections, but with the addition of a learnable sign-flipping function. The motivation for using this flip function comes from the findings presented in \cite{Ivan2020TrainingHE}. The gist of our approach is that new child filters can be created from parent weights through learning to flip signs in different parts of the parent filter block. This would imply that $\mathcal{F}_{c,i} = \mathcal{B}_i \odot \mathcal{F}_p$, where $\mathcal{B}_i \in \{0, 1\}^{k\times k \times m \times n}$ is the boolean mask matrix used to create the $\mathcal{F}_{c,i}$, and $\odot$ denotes the elementwise multiplication operation. A value of 1 denotes that the corresponding weight parameter in $\mathcal{F}_p$ needs to be multiplied by $-1$ when substituting in $\mathcal{F}_{c,i}$; 0 denotes otherwise. 
Note that although the size of this matrix is equal to $\mathcal{F}_{c,i}$ in terms of number of parameters, it stores only single-bit information which leads to drastic reduction in model size. To jointly learn sign-flippings along with the model parameters using gradient descent methods, we use the optimization paradigm as proposed in BiRealNet \cite{bireal}. 
% The sign-flippings are boolean masks and the added memory burden is negligible compared to the original model.

\section{Experiments}
\begin{table*}[]
\begin{center}
\begin{tabular}{cccccccc}
\toprule
\multicolumn{1}{l}{}           & \multicolumn{1}{l}{} & \multicolumn{2}{c}{\textbf{CIFAR-10}} & \multicolumn{2}{c}{\textbf{CIFAR-100}} & \multicolumn{2}{c}{\textbf{Tiny ImageNet}} \\
\cmidrule(l{2pt}r{2pt}){3-4} \cmidrule(l{2pt}r{2pt}){5-6} \cmidrule(l{2pt}r{2pt}){7-8}
\textbf{Method}                         & \textbf{Scale factor}         & \textbf{Acc. (\%)}  & \textbf{Rel. Cont. (\%)} & \textbf{Acc. (\%)}  & \textbf{Rel. Cont. (\%)}  & \textbf{Acc. (\%)}    & \textbf{Rel. Cont. (\%)}    \\
\midrule
\multirow{4}{*}{Dense}         & $1\times$                   & 91.19      & -               & 63.35      & -                & 47.37        & -                  \\
                               & $2\times$                   & 93.60      & -               & 72.45      & -                & 55.72        & -                  \\
                               & $4\times$                   & 94.91      & -               & 77.64      & -                & 59.94        & -                  \\
                               & $8\times$                   & 95.69      & -               & 80.19      & -                & 62.89        & -                  \\ \midrule
\multirow{3}{*}{Linear}        & $2\times$                   & 91.13      & -2.49           & 63.12      & -2.53            & 47.03        & -4.07              \\
                               & $4\times$                   & 91.50      & 8.33            & 63.33      & -0.14            & 46.35        & -8.11              \\
                               & $8\times$                   & 91.46      & 6.00            & 62.65      & -4.16            & 47.03        & -2.19              \\ \midrule
\multirow{3}{*}{$s$-RepeatNet} & $2\times$                   & 91.70      & 21.16           & 63.65      & 3.30             & 48.53        & 13.89              \\ 
                               & $4\times$                   & 91.72      & 14.25           & 65.42      & 14.49            & 49.55        & 17.34              \\
                               & $8\times$                   & 92.13      & 20.89           & 66.70      & 19.89            & 50.52        & 20.30              \\ \midrule
\multirow{3}{*}{$f$-RepeatNet} & $2\times$                   & 91.08      & 2.69            & 65.14      & 10.09            & 50.40        & 36.29              \\
                               & $4\times$                   & 92.67      & 39.78           & 70.81      & 52.20            & 56.80        & 75.02              \\
                               & $8\times$                   & 93.18      & 44.22           & 75.24      & 70.61            & 60.05        & 81.70              \\
\bottomrule
\end{tabular}
\end{center}
\caption{Performance scores obtained on the classification task of various datasets using the base ResNet-16 model, the scaled up versions obtained using RepeatNet as well as the dense counterparts. Dense networks are the full-precision fully trainable networks with same network architecture as the corresponding RepeatNet model. Here, `Rel. Cont.' denotes the fraction of improvement that the RepeatNet model obtains over the base model compared to the equivalent dense model.}
\label{tableres16}
\end{table*}
% 1. \% of performance improvement from physical size 
% 2. novel strategy to learn learnable transforms to get maximum improvement over previous. and \% performance improvement  

In this section, we conduct various sets of numerical experiments to demonstrate the contribution of model's physical size in its performance. The first set uses repeated width scaling to increase model's width while maintaining the parameters. The second set shows a similar setup but to assess the importance of depth in the model. Next, we also study the transferability of our scaled model architectures on the downstream task of semantic segmentation. We also show an example problem from the domain of group equivariant \mbox{CNNs \cite{cohen2016group,weiler2018learning}}, where our experiments reveal that the performance gains claimed to be obtained from inducing group-equivariance could partly be due to the repetition of weights in leading to its scale-up.

\subsection{Rescaling Network Architectures}

We demonstrate the efficacy of RepeatNet for depth and width scaling on VGG \cite{simonyan2014very} and ResNet \cite{he2016deep} models, respectively on CIFAR-10, CIFAR-100, Tiny-ImageNet and ImageNet datasets. All the models are trained using stochastic gradient descent method with an initial learning of 0.1, a momentum value of 0.9 and weight decay of 1e-3. For models trained on CIFAR-10 and CIFAR-100 datasets, we use 160 epochs and the learning rate is decayed by a factor of 0.1 after 80 and 120 epochs. Further, model are trained on Tiny-ImageNet and ImageNet datasets for 90 epochs and the learning rate is decayed by a factor of 0.1 after every 30 epochs.

\textbf{Width Scaling. }We study the influence of width scaling on two different ResNet models. Using RepeatNet, we scale the base ResNet models on CIFAR-10, CIFAR-100, TinyImageNet and ImageNet datasets. Architectures are made wider through repetitions while ensuring that the model size is still preserved. Table \ref{tableres16} reports performance scores for repetitions applied on ResNet-16 model for CIFAR and TinyImageNet datasets. We report scores for the base model (no scaling) as well as the various RepeatNet strategies discussed earlier. We scale the network by factors of $2\times$, $4\times$ and $8\times$. A general observation is that scaling the networks and making them wider improves its performance. Interestingly, we have observed on the TinyImageNet dataset that around 80\% of the performance gain can be achieved by $f$-RepeatNet at 8$\times$ scaling than what would have been obtained using the dense counterpart. For other cases as well, we see significant improvements at 8$\times$ scaling. An exception is the linear scaling approach where for some instances, it is observed that the performance drops slightly compared to the corresponding base model. 

% % Please add the following required packages to your document preamble:
% % \usepackage{graphicx}
% \begin{table}[h]
% \begin{center}

% \label{tab:resnet_imnet_table}
% \begin{tabular}{llcc}
% \toprule
% Rep. Factor & Tranf     & Rep Acc. (\%) & Rep. Cont. (\%) \\ \midrule
% \multicolumn{1}{c}{1}           & Dense     & 66.24         &                \\ \midrule
% \multicolumn{1}{c}{2}           & Dense     & 72.67         & -              \\
%             & Sign Flip & 69.96         & 57.85          \\ \midrule
% \multicolumn{1}{c}{4}           & Dense     & 76.60          & -              \\
%             & Sign Flip & 72.41         & 59.56         \\ \bottomrule
% \end{tabular}%
% \end{center}
% \caption{}
% \label{tableres18}
% \end{table}

% \usepackage{multirow}
\begin{table}[]
\begin{center}
\scalebox{0.964}{
\begin{tabular}{cccc}
\toprule
\textbf{Method}                         & \textbf{Scale factor}      & \textbf{Acc (\%)} & \textbf{Rel. Cont. (\%)} \\\midrule
\multirow{3}{*}{Dense}         & 1$\times$           & 66.24    & -               \\
                               & 2$\times$           & 72.67    & -               \\
                               & 4$\times$           & 76.6     & -               \\\midrule
\multirow{2}{*}{$f$-RepeatNet} & 2$\times$           & 69.96    & 57.85           \\
                               & 4$\times$           & 72.41    & 59.56           \\\bottomrule
\end{tabular}
}
\end{center}
\caption{Performance scores obtained on the task of classification of Imagenet samples obtained using ResNet-18, its RepeatNet variants as well as the corresponding dense scaled-up models. Here, `Rel. Cont.' denotes the relative contribution of RepeatNet and can be interpreted as the gain obtained by using RepeatNet normalized by the total gain observed by the corresponding dense architecture.}
\label{tableres18}
\end{table}

The efficacy of our RepeatNet strategy is better demonstrated through applicability on the large-scale \mbox{ImageNet-1k} dataset. Table \ref{tableres18} shows the performance values for the ResNet-18 architecture obtained on this dataset. Similar to the previous experiments, we observe that models made wider using RepeatNet delivered significantly improved performance. The RepeatNet models are able to improve the performance of the base model (1$\times$) by approximately 60\% of what could be achieved by scaling as a dense model. 
% We observe that width scaling with RepeatNet improves the performance of all ResNet models for the two datasets consistently.

% and \ref{tableres18} report results for width scaling on CIFAR10, CIFAR100, Tiny-ImageNet and ImageNet datasets, respectively. We observe that width scaling with RepeatNet improves the performance of all ResNet models for the two datasets consistently.

% \begin{figure}[!b]
% \centering
% {\includegraphics[scale=0.4]{results/vgg_tiny.png}};
% \caption{Performance scores for Base VGG (4-8) and RepeatNet on TinyImageNet for different choices of number of layers.Here, all the models are scaled upto VGG11.}
% \label{fig_schem1}
% \end{figure}
% \begin{figure}[!b]
% \centering
% {\includegraphics[scale=0.4]{results/vgg_c10.png}};
% \caption{Performance scores for Base VGG (4-8) and RepeatNet on CIFAR-10 for different choices of number of layers. Here, all the models are scaled upto VGG11.}
% \label{fig_schem1}
% \end{figure}
% \begin{figure}[!b]
% \centering
% {\includegraphics[scale=0.4]{results/vgg_c100.png}};
% \caption{Performance scores for Base VGG (4-8) and RepeatNet on CIFAR-100 for different choices of number of layers. Here, all the RepeatNet models are scaled upto VGG11.}
% \label{fig_schem1}
% \end{figure}

\begin{figure*}
\begin{center}
\begin{subfigure}{0.33\textwidth}
  \includegraphics[width=\linewidth]{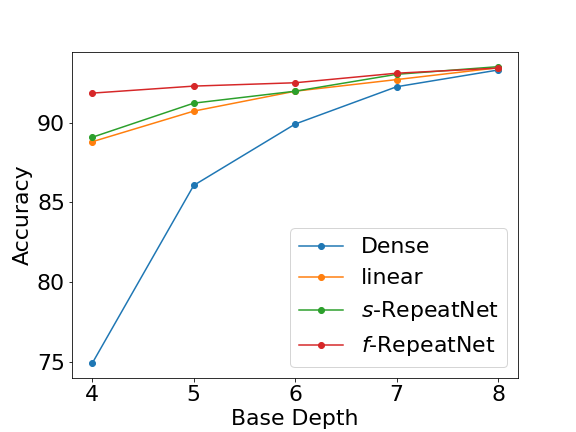}
  \caption{CIFAR-10}
  \label{fig:ad}
\end{subfigure}
\begin{subfigure}{0.33\textwidth}
  \includegraphics[width=\linewidth]{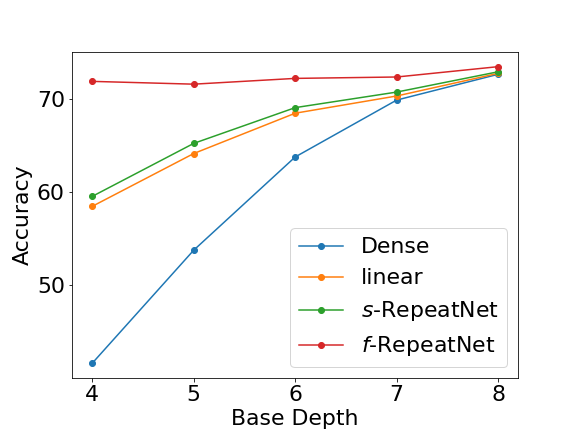}
  \caption{CIFAR-100}
  \label{fig:product_page}
\end{subfigure}
\begin{subfigure}{0.33\textwidth}
  \includegraphics[width=\linewidth]{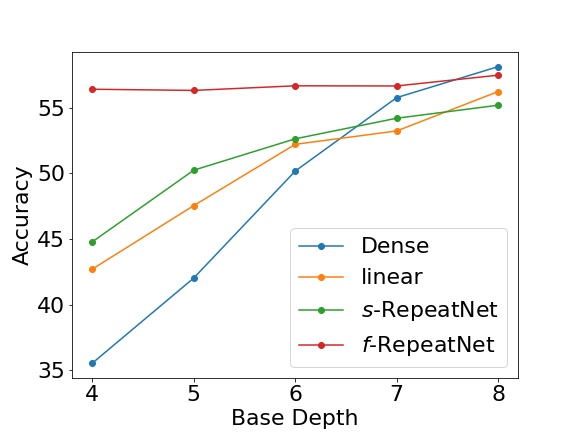}
  \caption{Tiny-ImageNet}
  \label{fig:product_page}
\end{subfigure}
\caption{Performance scores for Base VGG (4-8) and RepeatNet on CIFAR-100 (a), CIFAR-100 (b) and Tiny-ImageNet (c) for different choices of number of layers. Here, all the RepeatNet models are scaled upto VGG11.}
\label{fig_vgg_results}
\end{center}
\end{figure*}

% We took ResNet16 for CIFAR10, CIFAR100, Tiny-ImageNet and ResNet 18 for ImageNet as the base architecture and used its width scaled variants as the baseline, We used our filter repetition and filter transformation techniques to match the physical size of base model to width scaled networks while keeping the number of parameters fixed. The results are shown in Table \ref{}. 

\textbf{Depth Scaling. }We analyze here the applicability of RepeatNet for depthwise scaling of CNN networks. For depth scaling, we take VGG\footnote{Unlike the standard VGG networks, our models have only one fully-connected layer at the end.} models of different sizes spanning from VGG4 to VGG8.  Further, depth scaling is performed using RepeatNet strategy on each of our VGG models such that their sizes become equivalent to that of VGG11. Performance scores for the VGG models and the respective RepeatNet versions are reported in Fig. \ref{fig_vgg_results}. It is observed that depth scaling with RepeatNet strategy can significantly improve the model performance on CIFAR10, CIFAR100 and Tiny-ImageNet datasets. Note that repetitions in smaller models are more than in bigger models, thus showing higher relative performance gain in the smaller ones.

% To show the importance of depth in model's performance we perform multiple experiments on VGG\cite{}, with CIFAR10, CIFAR100 and Tiny-ImageNet datasets. While repeating a VGG model with n original layers and m repeated layers, weights from ni layer are repeated into mj layer where j\%i==0. Assuming Repetition from 3-layer network to a 8-layer network with weights w1, w2 and w3 of respective conv layers of 3-layer network, the weights of VGG3 can be repeated to match layerwise shapes in the following order - [w1, w2, w3, w1, w2, w3, w1, w2].

\begin{figure*}
\begin{center}
	\includegraphics[width=0.9\linewidth]{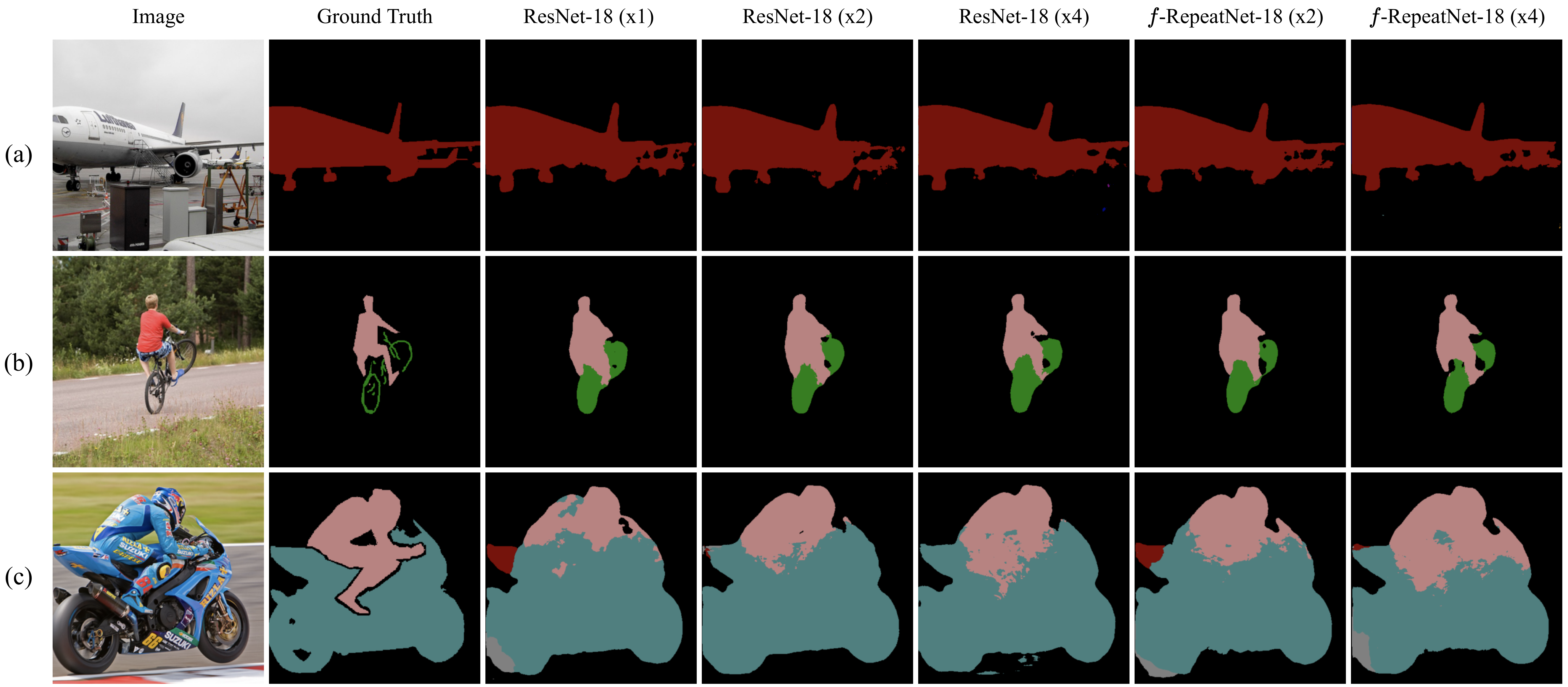}
\end{center}
\caption{Example image samples and the the corresponding ground-truth segmentation masks from Pascal-VOC dataset as well as the masks predicted using ResNet-18 and its scaled variants obtained with RepeatNet. }
\label{fig:pascalvoc}
\end{figure*}

\begin{table}[]

\label{tab:pascal_table}
\begin{center}
\begin{tabular}{lccc}
\toprule
\textbf{Method}         & \textbf{Scale factor} & \multicolumn{1}{l}{\textbf{mIOU (\%)}} & \textbf{Acc. (\%)} \\ \midrule
                        & 1$\times$             & 59.5                                   & 89.4               \\
Dense                   & 2$\times$             & 64.7                                   & 91.0               \\
                        & 4$\times$             & 68.0                                   & 91.8               \\ \midrule
\multirow{2}{*}{$f$-RepeatNet}& 2$\times$           & 62.1                                   & 90.2               \\ 
                          & 4$\times$           & 64.5                                   & 90.7               \\\bottomrule
\end{tabular}
\end{center}
\caption{Performance scores for the task of segmentation on Pascal-VOC dataset obtained using encoders from the base ResNet-18 models, its scaled up variants obtained with RepeatNet as well as the corresponding dense counterparts. Here, the memory requirements of RepeatNet models are the same as the base model, while those for the dense models scale by the corresponding `Scale factor' term.}
\label{tab:pascal_table}
\end{table}
\subsection{Rescaling Transferability}
Due to the lack of enough training samples on a downstream task, and the limited computational budget available, model weights trained on one task are often used as initialization for some other task. For example, various segmentation models use encoders that are extracted from the backbone of classification models trained on ImageNet or other large-scale datasets. We study here whether the architectures obtained after scaling with RepeatNet are also transferable to other downstream tasks. We train ResNet-18 model for the classification task on ImageNet-1k and use the backbone as an encoder for a segmentation model. Segmentation performance is evaluated on the Pascal-VOC dataset. We perform 2$\times$ and 4$\times$ width scaling of the base architecture and compare with the respective dense counterpart. For all the cases, a common U-Net \cite{Ronneberger2015UNetCN} architecture with a 4 layer decoder is used. We apply 1x1 convolutions as skip connections between encoder and decoder to map wide activations of the encoder to their normal counterparts as input to the decoder. This allows for a fair comparison as all the models have exactly similar decoder network.

Table \ref{tab:pascal_table} shows the results of segmentation performance obtained using the base ResNet-18 model as well as the RepeatNet variants. To highlight the significance of the performance gains obtained with RepeatNet, we also report scores obtained with the dense architectures for every scaled up architecture obtained with RepeatNet. We see that Repeatnet architectures obtained with 2$\times$ and 4$\times$ scaling improve the mean IoU scores of the base model by 2.6\% and 5\%, respectively. This accounts for 50\% and 58.8\% of the total improvements obtained with the corresponding dense architectures. However, it is important to note that while the model size of the RepeatNet architectures is approximately the same as the base model, it becomes $\sim$ 4$\times$ and  $\sim$ 16$\times$ for the respective 2$\times$ and 4$\times$ scaled dense architectures. Clearly, the improvements reported in Table \ref{tab:pascal_table} show that the RepeatNet architectures are transferable across tasks. 

For qualitative analysis, we also show in Fig. \ref{fig:pascalvoc} a few randomly chosen image samples from Pascal-VOC dataset. For these samples, we show the ground-truth masks, those predicted using ResNet-18 model, results of RepeatNet-scaled variants as well as those with the dense counterparts. First observation is that the segmentation results improve for the scaled up networks, be it the RepeatNet models or the dense ones. Moverover, the results of RepeatNet seem visually similar to those obtained with the dense networks at same scale.

\begin{table}[h]
\begin{center}
\scalebox{0.92}{
\begin{tabular}{ccccc}
\toprule
\multicolumn{2}{c}{\textbf{Rot-Equivariant CNN}} & \multicolumn{2}{c}{\textbf{$s$-RepeatNet}} &  \\ \cmidrule(l{2pt}r{2pt}){1-2} \cmidrule(l{2pt}r{2pt}){3-4}
\textbf{Group} & \textbf{Acc.} (\%) & \textbf{Used} & \textbf{Acc.(\%)} & \textbf{Rel. Cont. (\%)}\\ \midrule
- & - & \xmark & 92.6 & - \\
 C4 & 96.5 & \cmark & 93.2 & \textbf{15.4}\\
 C8 & 96.8 & \cmark  & 94.4 & \textbf{42.9}\\
C16 & 96.3 & \cmark  & 93.5 & \textbf{24.3}\\
%& 4 & 183498 & 0.9225 \\ % 120
%& 8 & 194042 & 0.9269 \\ % 256
%& 10 & 199626 & 0.9265 \\ % 330
\bottomrule
\end{tabular}
}
\end{center}
\caption{Quantitative assessment of Relative Contribution (Rel. Cont.) in accuracy observed purely due to learned repetition of weights presented as a percentage of overall gain observed when using rotation-equivariant CNNs instead of vanilla CNNs for the task of classification of the rot-MNIST digits. Note that the RepeatNet models are chosen such that the inference FLOPs are approximately similar to the respective RE-CNN model. The model denoted with `\xmark' indicates the vanilla CNN model with number of training parameters equal to the RE-CNN models and is used as a baseline to evaluate the performance gains for other models.}
\label{tablegcnn}
\end{table}

\subsection{Role in Group-Equivariant Networks}
In the experiments above, we have shown that it is possible to improve the performance of any existing CNN model through scaling up its architecture using RepeatNet. This also implies that for deep learning problems where sharing of weights occurs, it is important that network scaling is also taken into consideration while deducing conclusions. In this regard, we consider here an example problem from the domain of group-equivariant CNNs. We particularly focus on Rotation-Equivariant Steerable CNN, referred further as RE-CNN. This approach induces rotation-equivariance into the CNN architecture through restricting the choice of weights of the network to be build from the combination of a certain class of basis functions, \emph{e.g.}, Gaussian radial basis.

We conduct here an experiment with RE-CNN to analyze whether the gain in performance observed on rot-MNIST dataset is purely due to the inbuilt property of rotation equivariance of the model, or whether part of it is due to repetition of weights. Note that while we particularly focus on rotation-equivariant models in this paper, the idea holds for other transformation groups as well. For RE-CNN, we use steerable filter CNN 
\footnote{Rotation equivariant CNN is implemented in this paper using the \texttt{e2cnn} python library available at \texttt{https://github.com/QUVA-Lab/e2cnn}.} implementation as described in \cite{e2cnn} and experiment with rotation groups of 4, 8 and 16. The choice of model architecture is the same as described in \cite{e2cnn}, except that the number of channels in each convolutional layer are reduced by a factor of 2. 

Performance scores for these models are reported in Table \ref{tablegcnn}. Further, we use the baseline vanilla CNN model with same number of optimization parameters and upscale its size to create three $s$-RepeatNet variants that are equivalent to the three RE-CNN models in terms of the inference FLOPs. Training as well as testing are performed on rot-MNIST dataset. For the training of all models, we follow the training strategy as described in \cite{e2cnn}.

From Table \ref{tablegcnn}, we see that both, RE-CNN as well as RepeatNet models improve the classification performance compared to the baseline. Interestingly, with RepeatNet scaling, up to 42.9\% of the gain is already observed than what is obtained with RE-CNN. Given that weight sharing already improves the performance, it is implied that part of the performance gain in RE-CNN could actually be due to the implicit weight sharing and the resultant scaling involved in its formulation. Clearly, it would be unfair to make a comparison between RE-CNN and vanilla CNN at same number of optimization parameters. We argue that any such comparison should only be performed at fixed number of FLOPs.

\section{Discussion}

In this paper, we presented \emph{RepeatNet}, a novel strategy to scale CNNs and make them deeper and wider without adding extra optimization parameters to the model. We demonstrated through numerical experiments that our strategy of scaling models using learnable nonlinear transformation functions can significantly improve the performance of the existing models. The results presented in the study reveal that performance gained from scaled networks is only partly due to the addition of extra parameters, and part of it comes from the increased physical size of the model itself. The relevance of these insights is further supported from the observations on group-equivariant CNNs, where we showed that upto 40\% of the performance gain credited to the implicit property of group equivariance in the model, could actually be due to just weight sharing. 

\begin{table}[]
\begin{center}
\begin{tabular}{lccc}
\toprule
\textbf{Method} & \textbf{Scale factor} & \textbf{Mean Acc.} & \textbf{Std Dev}\\ \midrule

Dense & 1$\times$ & 46.61 & 0.37 \\ \midrule

\multirow{2}{*}{Linear} & 4$\times$             & 47.21                                  & 0.27              \\
                        & 8$\times$             & 47.03                                   & 0.34               \\ \midrule
\multirow{2}{*}{$s$-RepeatNet} & 4$\times$             & 49.48                               & 0.80              \\
                        & 8$\times$             & 51.49                                   & 0.50               \\ \midrule
\multirow{2}{*}{$f$-RepeatNet}& 4$\times$           & 57.22                                   & 0.34             \\ 
                          & 8$\times$           & 60.14                                 & 0.45 \\ \bottomrule
\end{tabular}
\end{center}
\caption{Mean accuracies (in \%) and standard deviation scores computed over 10 runs for the base ResNet-16 model and its RepeatNet variants on TinyImageNet dataset.}
\label{tab:sdtable}
\end{table}

While we have reported massive absolute improvements with network scaling, the significance of these results can only be validated based on the error observed in the performance across multiple runs. In this regard, we performed 10 runs for several variants of RepeatNet for ResNet-16 model on the Tiny-ImageNet dataset. Table \ref{tab:sdtable} reports the details of the different runs including mean accuracy score and the deviation. We observe that the deviations in mean accuracy scores are very small for the base model (1$\times$) as well as its RepeatNet variants. The maximum is reported for $s$-RepeatNet with a standard deviation score of 0.80. However, compared to the difference in accuracy scores for the base model and the RepeatNet variants, it is clear that the deviation is small, especially for our $s$-RepeatNet and $f$-RepeatNet models. This demonstrates that the \mbox{improvements} gained by RepeatNet are significant. In future, it would be of interest to conduct a similar study on other experiments as well.

\textbf{Limitations. }We have shown that scaling networks \mbox{using} RepeatNet leads to improved model performance. However, there are are several limitations as well. First, there is still some gap in performance scores between RepeatNet and the respective dense implementations. We believe that with more improved nonlinear projection strategies, this gap can be narrowed down further. We further speculate that a hybrid of depth and width scaling might work better that both of these individually, however, this has not been explored, and it could be a future direction to explore further.
\small
\bibliographystyle{ieee_fullname}
\bibliography{egbib}

\end{document}